\documentclass[sigconf]{acmart}

\AtBeginDocument{%
  \providecommand\BibTeX{{%
    \normalfont B\kern-0.5em{\scshape i\kern-0.25em b}\kern-0.8em\TeX}}}

\setcopyright{acmcopyright}
\copyrightyear{2023}
\acmYear{2023}
\acmDOI{10.1145/1122445.1122456}

\acmConference[KDD '23]{KDD '23: Knowledge Discovery and Data Mining}{August 6--8, 2023}{Long Beach, CA, USA}
\acmBooktitle{KDD '23: Knowledge Discovery and Data Mining,
  August 6--8, 2023, Long Beach, CA, USA}
\acmPrice{15.00}
\acmISBN{978-1-4503-XXXX-X/18/06}

\usepackage{graphicx}
\usepackage{tabularx}
\usepackage{enumerate}
\usepackage{multirow}
\usepackage{listings}
\usepackage{xcolor}
\usepackage{comment}
\usepackage{caption}
\definecolor{codegreen}{rgb}{0,0.6,0}
\definecolor{codegray}{rgb}{0.5,0.5,0.5}
\definecolor{codepurple}{rgb}{0.58,0,0.82}
\definecolor{backcolour}{rgb}{0.95,0.95,0.92}

\lstdefinestyle{mystyle}{
	backgroundcolor=\color{backcolour},   
	commentstyle=\color{codegreen},
	keywordstyle=\color{magenta},
	numberstyle=\tiny\color{codegray},
	stringstyle=\color{codepurple},
	basicstyle=\ttfamily\footnotesize,
	breakatwhitespace=false,         
	breaklines=true,                 
	captionpos=b,                    
	keepspaces=true,                 
	numbers=left,                    
	numbersep=5pt,                  
	showspaces=false,                
	showstringspaces=false,
	showtabs=false,                  
	tabsize=2
}

\begin{document}

\title{Unlocking Sales Growth: Account Prioritization Engine with Explainable AI}

\author{Suvendu Jena}
\affiliation{\institution{LinkedIn Corporation}\city{Sunnyvale}\state{CA}\country{USA}}
\email{sjena@linkedin.com}

\author{Jilei Yang}
\affiliation{\institution{LinkedIn Corporation}\city{Sunnyvale}\state{CA}\country{USA}}
\email{jlyang@linkedin.com}

\author{Fangfang Tan}
\affiliation{\institution{LinkedIn Corporation}\city{Sunnyvale}\state{CA}\country{USA}}
\email{ftan@linkedin.com}

\renewcommand{\shortauthors}{Suvendu Jena, Jilei Yang and Fangfang Tan}

\begin{abstract}
B2B sales requires effective prediction of customer growth, identification of upsell potential, and mitigation of churn risks. LinkedIn sales representatives traditionally relied on intuition and fragmented data signals to assess customer performance. This resulted in significant time investment in data understanding as well as strategy formulation and under-investment in active selling. To overcome this challenge, we developed a data product called Account Prioritizer, an intelligent sales account prioritization engine. It uses machine learning recommendation models and integrated account-level explanation algorithms within the sales CRM to automate the manual process of sales book prioritization. A successful A/B test demonstrated that the Account Prioritizer generated a substantial +8.08\% increase in renewal bookings for the LinkedIn Business.
\end{abstract}



\maketitle

\section{Introduction}
At LinkedIn, we have two broad categories of sales representatives (reps): Account Executives (AEs) who focus on acquiring new customers and converting them into first-time LinkedIn B2B buyers, and Account Directors (ADs) who serve existing customers and focus on their growth. Our focus is ADs and optimizing their books: At the beginning of every fiscal year, ADs are assigned a book consisting of a set of accounts.\footnote{In the context of this paper, the terms "Accounts" and "Customers" are interchangeable and refer to the same entities.} Their responsibility is to renew these accounts and drive growth by upselling and cross-selling LinkedIn's hiring and learning products, following the SaaS renewal cycle. Depending on the size of the accounts, an AD can have anywhere from 5 to 60+ accounts in their book. Prioritizing this book becomes a fundamental challenge as all accounts must be serviced during the renewal cycle, and quarterly/monthly sales planning involves identifying accounts with growth potential versus those at risk.

Historically, ADs relied on multiple isolated dashboards and their field intuition to gather and understand whether an account is likely to grow or churn. This approach resulted in ADs spending extensive time crunching data, which could be potentially used to do active selling. Therefore, there was a need for an automated prioritization engine. In this paper, we describe the development of Account Prioritizer, a data product at LinkedIn that predicts upsell opportunities and churn risks overall \& across the constituent LinkedIn hiring and learning products (such as Recruiter, Jobs, etc.), providing ADs the time to execute sales strategies, convert these opportunities into actual sales outcomes and mitigate churn risks.

Additionally, Account Prioritizer, in conjunction with CrystalCandle \cite{yang2021crystalcandle} explanation engine provides instance level natural language explanations for these predictions, which helps ADs understand the why behind the recommendations and guides their account outreach efforts. The explanation layer also increases trust in the system and leads to higher adoption among ADs. Finally, we discuss how we leverage experiments and observational causal studies to measure the incremental impact of these models throughout their lifecycle. By continuously evaluating our approach, we ensure the effectiveness and ongoing improvement of Account Prioritizer in supporting ADs' decision-making processes and driving sales success.

\section{Methodology}

\subsection{Problem Formulation}
Let's say we have accounts $a_1, a_2, \ldots, a_n$ in the book of a sales representative $r$. These accounts have in time $T-1$ (previous renewal cycle) spent amount $ps_1, ps_2, \ldots, ps_n$ with LinkedIn. Let's denote the spend that these accounts will do with LinkedIn in the upcoming renewal cycle $T$ as $cs_1, cs_2, \ldots, cs_n$. We can have the following outcomes for account $a_i$ in renewal cycle $T$ ($i=1,\ldots,n$):
\begin{enumerate}
    \item 
    $cs_i>ps_i \rightarrow$ Upsell of $(cs_i-ps_i)$.
    \item 
    $cs_i=ps_i \rightarrow$ Stay flat since $(cs_i-ps_i)=0$.
    \item 
    $cs_i<ps_i \rightarrow$ Churn of $(cs_i-ps_i)$.
\end{enumerate}
Since the values of $ps_1, ps_2, \ldots, ps_n$ are known, we need to predict $(cs_i-ps_i)$. Once we get the values of $(cs_i-ps_i)$ for all accounts $a_1, a_2, \ldots, a_n$, we can then use it to prioritize the book of sales rep $r$ as an ordered list of
$$(cs_{s_1}-ps_{s_1}), (cs_{s_2}-ps_{s_2}), \ldots, (cs_{s_n}-ps_{s_n})$$
from highest to lowest. The larger positive values denote higher upsell potential, smaller negative values denote higher churn risk, and values near 0 denote accounts staying relatively flat.

In addition to analyzing the overall spend at the account level, we offer sales reps valuable insights into upselling opportunities and potential risks associated with specific products. This becomes particularly relevant when customers engage with multiple LinkedIn products. To achieve this, we can formulate a similar approach with previous product quantity $pq_1, pq_2, \ldots, pq_n$ and upcoming product quantity $cq_1, cq_2, \ldots, cq_n$ to make an ordered list of
$$(cq_{q_1}-pq_{q_1}), (cq_{q_2}-pq_{q_2}), \ldots, (cq_{q_n}-pq_{q_n})$$
for each LinkedIn product.

\subsection{Label Creation}

\subsubsection{Label Granularity}

\begin{figure}
\centering
  \includegraphics[width=\columnwidth]{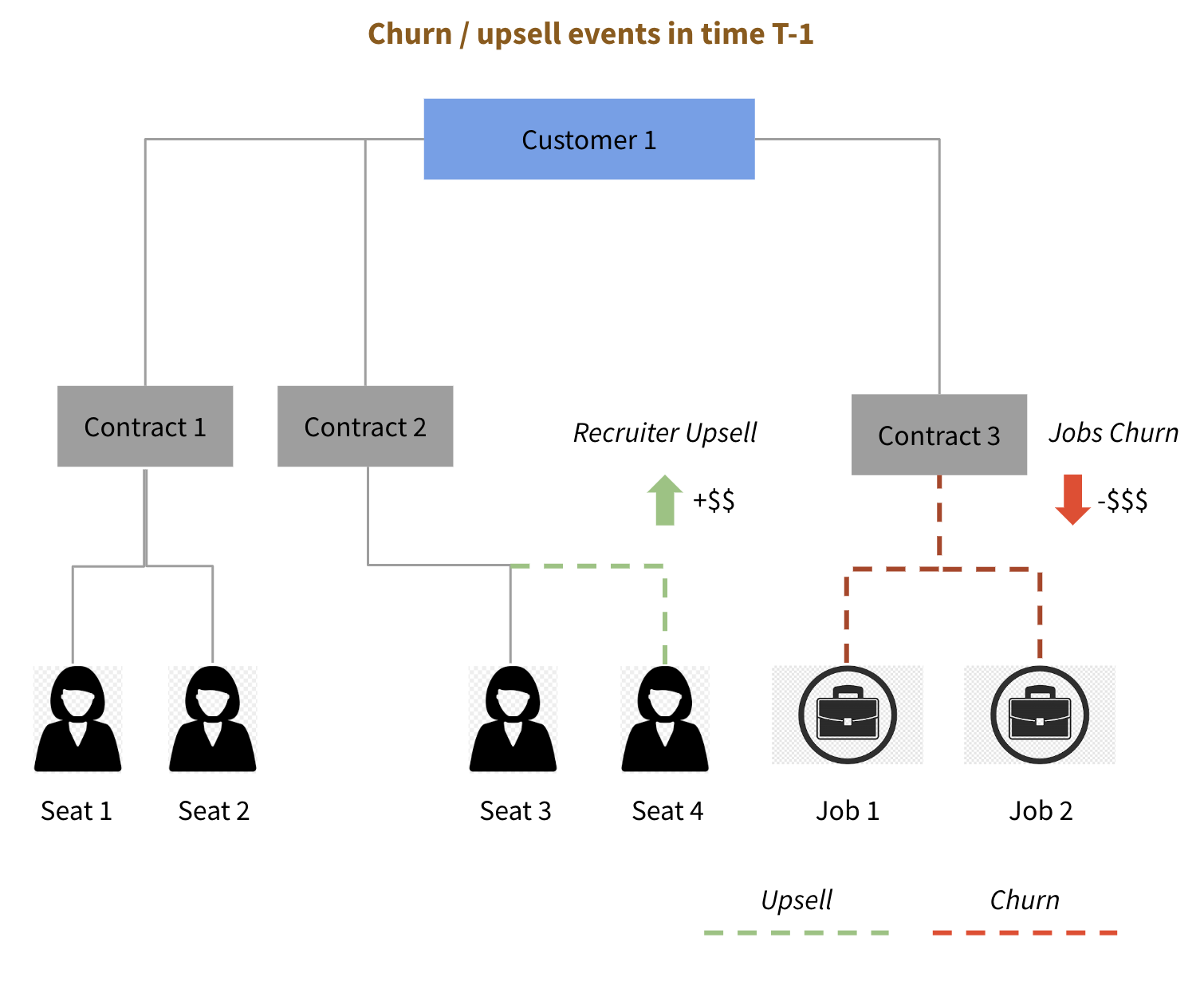}
  \caption{Historical upsell/churn events.}~\label{fig:historical_upsell_churn_events}
\end{figure}

\begin{figure}
\centering
  \includegraphics[width=\columnwidth]{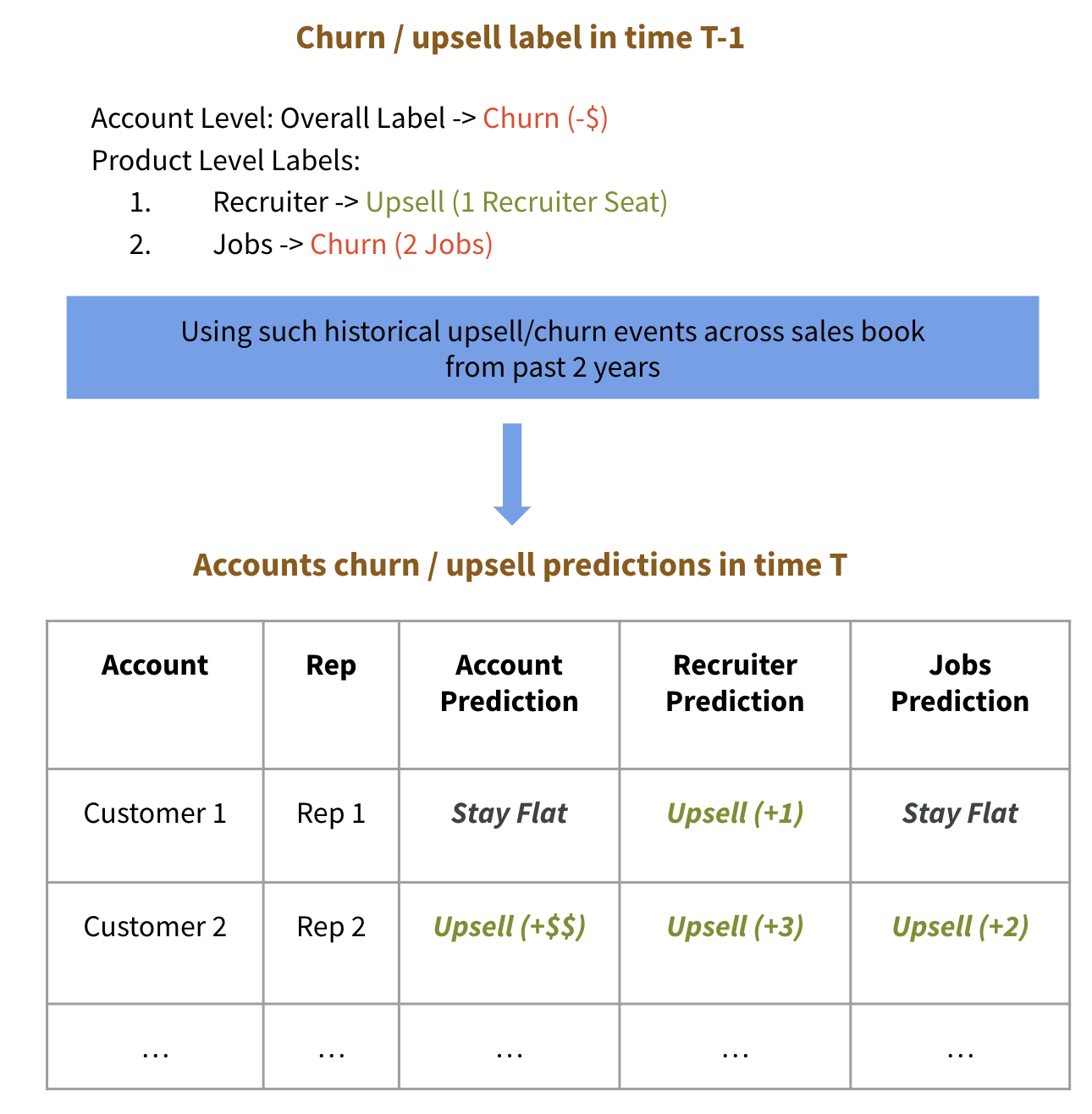}
  \caption{Upsell/Churn labels \& predictions.}~\label{fig:upsell_churn_labels _predictions}
\end{figure}

An account serviced by the sales reps could purchase multiple types of LinkedIn products. The customer's spending behavior varies based on the product type and quantity. Consequently, there are numerous upsell and churn events that occur for each account. Here is an example to illustrate this:
\begin{enumerate}
    \item 
    As shown in Figure \ref{fig:historical_upsell_churn_events}, consider Customer 1 who has 3 ongoing contracts with LinkedIn. They increased the number of Recruiter Licenses\footnote{LinkedIn recruiter license is recruiting tool to source, contact, and hire the right candidates faster.} by adding one additional Recruiter License to Contract 2. Simultaneously they also reduced their spend on Jobs\footnote{LinkedIn jobs platform helps companies to post their jobs on LinkedIn and easily target, prioritize, and manage qualified applicants.} by removing two job licenses from Contract 3. Since the additional spend on Recruiter Licenses (upsell) is outweighed by the reduction in spend on Jobs (churn), the overall spending of Customer 1 with LinkedIn decreases at time $T-1$.
    \item 
    As highlighted in Figure \ref{fig:upsell_churn_labels _predictions}, this scenario leads to the creation of three distinct labels at time $T-1$ (one overall \& one for each of the two products) . At the overall account level, we observe a churn label indicating a reduction in overall account spending (given the individual product spend addition was less than spend on products removed). At the individual product level, we have an upsell label for the Recruiter product and a churn label for the Jobs product.
\end{enumerate}

By collecting these upsell and churn labels gathered from the accounts (both overall spend, and products added/removed) over the past two years, we aim to predict the upsell and churn potential for all the accounts due for renewal in the upcoming time period $T$.

Through the prediction of both overall account-level spend and constituent product-level quantities, our approach enables the provision of granular and tailored recommendations to sales reps. This empowers them to strategically prioritize accounts based on their potential for higher overall spend, while also facilitating informed decisions regarding the optimal product mix for each individual customer.

\subsubsection{Capturing upsell/churn events throughout the SaaS renewal cycle}
\begin{figure}
\centering
  \includegraphics[width=\columnwidth]{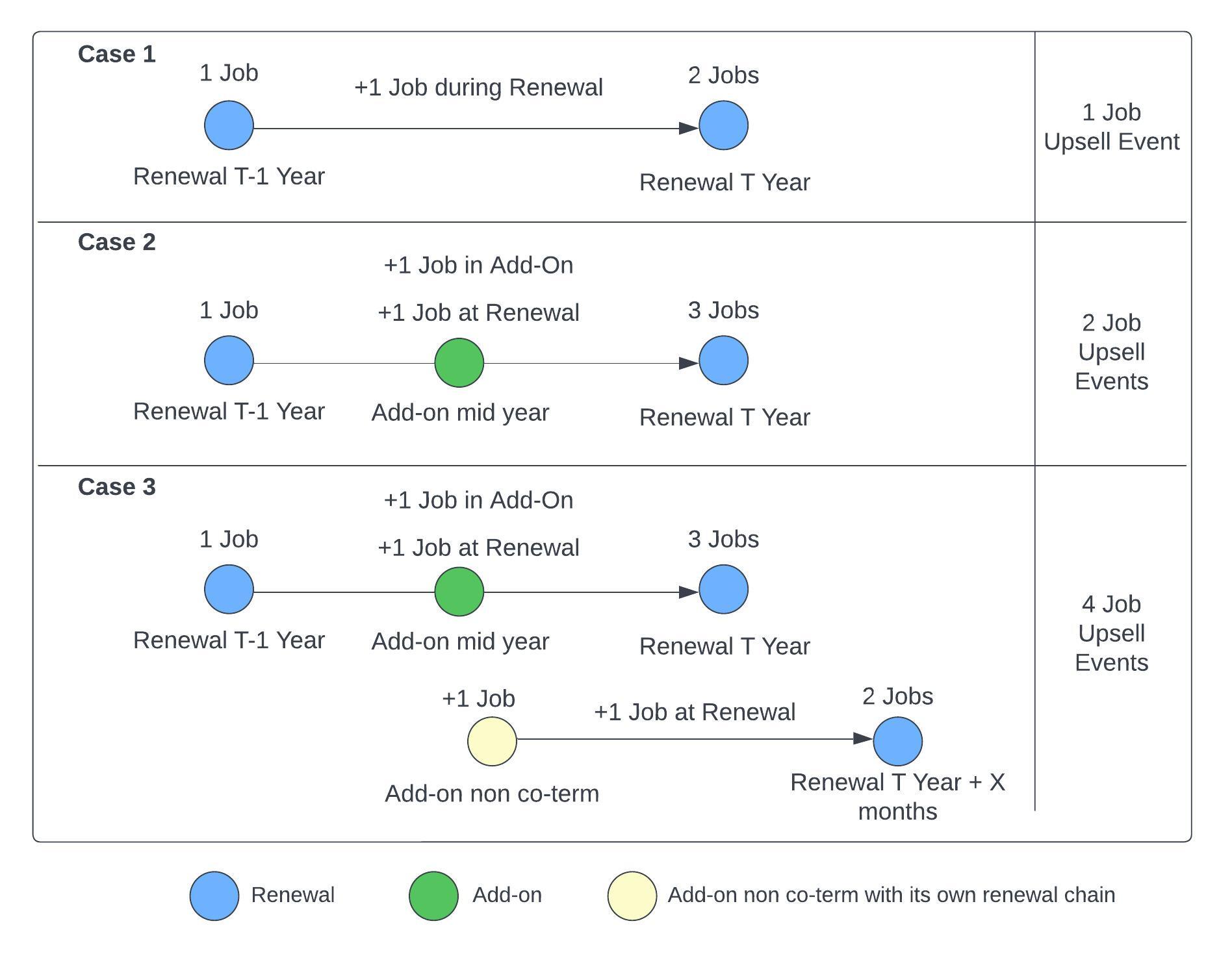}
  \caption{Different scenarios of upsell events.}~\label{fig:different_scenarios_upsell_events}
\end{figure}

As with other SaaS products, while churn happens during renewal, upsell can happen throughout the year - e.g., customer increased spend mid-cycle by adding additional products to their contract. This leads to a waterfall trend of product bookings/quantity throughout the renewal cycle, so defining a label for churn/upsell becomes challenging. To illustrate this, let's consider Figure \ref{fig:different_scenarios_upsell_events}.
\begin{enumerate}
    \item 
    In Case 1, we have a straightforward scenario where a customer adds an extra Jobs License during renewal in time $T-1$, resulting in a single upsell event. This case has already been covered in the previous section.
    \item 
    In Case 2, in addition to Case 1’s 1 upsell event, there is an additional upsell event since the customer added an extra Jobs License mid-cycle before the renewal (through an add-on opportunity\footnote{Add-on opportunity is a mid renewal cycle customer purchase event, wherein the customer adds more products to their portfolio and the opportunity renews as per the regular renewal cycle.}), making a total of two recorded upsell events.
    \item 
    Finally, in Case 3, we encounter a non-renewal cycle add-on opportunity called add-on non-co term\footnote{Add-on non co-term opportunity is similar to a regular add-on opportunity except that it triggers its own renewal cycle.} (non co-term opportunities differ from regular add-on because non-co term triggers their own renewal chain). The customer on top of the two upsell events in Case 2, adds one more Jobs License during add-on non co-term opportunity and a 4th Jobs License during the non co-term renewal. In totality, this scenario results in 4 upsell events for this customer.
\end{enumerate}

\begin{figure}
\centering
  \includegraphics[width=\columnwidth]{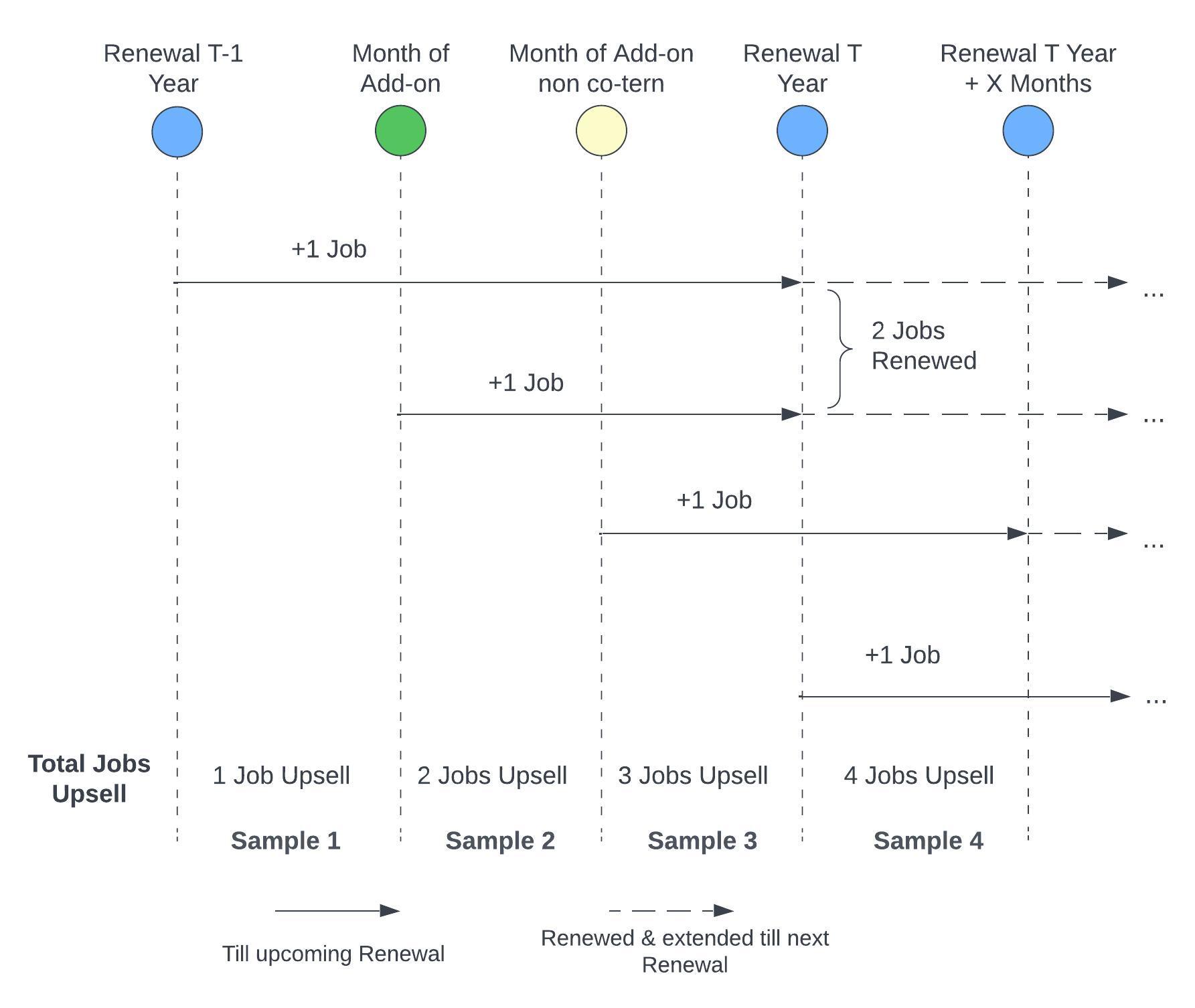}
  \caption{Label generation across overlapping time periods for Case 3.}~\label{fig:label_generation}
\end{figure}

In order to collect all these upsell events across the renewal cycle, we implement a system where labels are generated in overlapping time periods throughout the cycle, instead of assigning a single label to an entire renewal cycle. In order to explain this, we take the example of Case 3 in Figure \ref{fig:different_scenarios_upsell_events}, and illustrate how labels are collected in this specific case, as illustrated in Figure \ref{fig:label_generation}:
\begin{enumerate}
    \item 
    Collect Sample 1: 1st Jobs upsell event is captured when we collect the label from Renewal $T-1$ Year to Month of add-on, with the total upsell quantity = 1 Job.
    \item 
    Collect Sample 2: Between Month of add-on and Month of add-on non co-term we capture the 2nd Jobs upsell event, taking the total upsell quantity = 2 Jobs.
    \item 
    Collect Sample 3: Between Month of add-on non co-term and Renewal $T$ Year, we capture the 3rd Jobs upsell event, with the total upsell quantity = 3 Jobs.
    \item 
    Collect Sample 4: Finally, between Renewal $T$ Year to Renewal of the non co-term, we capture the 4th Jobs upsell event, taking the total upsell quantity = 4 Jobs.
\end{enumerate}

Hence for Case 3, we have 4 samples in the training dataset mapping to 4 Job upsell events. On the similar lines, for Case 1 we will have a single sample in the training dataset, and 2 samples for Case 2. Although this approach increases the complexity of the features used for labeling and generates multiple samples for each account, it provides more accurate labeling and a larger sample size across various customers.

\subsection{Feature Generation}

\subsubsection{Features used}
In the feature selection process, we curated a set of informative features to capture the key aspects influencing the upsell/churn prediction task. Some of the most important feature categories included past purchasing patterns, product usage, delivered ROI from LinkedIn to customers, spend in other LinkedIn business lines, online purchases of LinkedIn subscription products, talent trends such as hiring \& attrition, customer traffic on LinkedIn.com as well as various macro features. We observe that product usage and ROI continue to be significant factors in a customer deciding to buy more.

\begin{figure*}[t]
\centering
  \includegraphics[width=2\columnwidth]{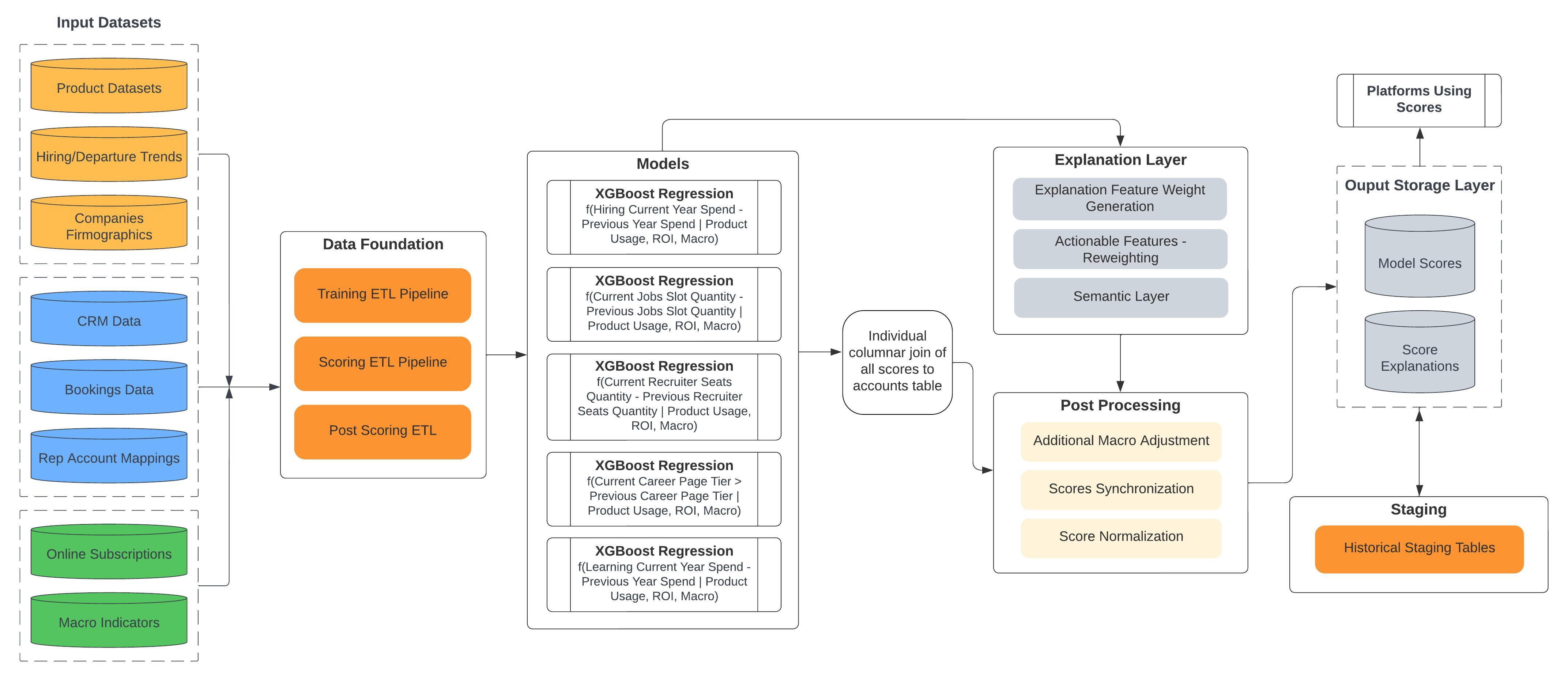}
  \caption{Model architecture and deployment.}~\label{fig:model_architecture_and_deployment}
\end{figure*}

\subsubsection{Feature preprocessing}
In the data preprocessing phase, we employed several techniques to prepare the dataset for our machine learning model, aiming to enhance feature representation and address specific challenges in the data. Through an extensive exploratory data analysis (EDA), we identified highly correlated features and applied transformations to derive rate features. This process involved normalizing the features by dividing them with a reference variable for calculating ratios, effectively capturing the relative proportions or rates between the variables. For example we calculate the acceptance rate of recruiter messages by candidates rather than looking at only the volume of messages sent by recruiters. By incorporating these rate features, we mitigated the issue of multicollinearity, which can hinder model interpretability and potentially lead to overfitting.

To handle categorical features, we employed label response encoding, a technique that maps categories to their corresponding target variable values. This approach not only converted categorical variables into numerical representations but also preserved the underlying relationship between the categories and the target variable. Label response encoding enables the model to effectively capture the impact of categorical features on the prediction task. Furthermore, to address outliers and extreme values in the Year-over-Year (YoY) booking features, we applied a capping mechanism. By setting upper limits on these variables, we restricted the influence of outliers on the model's training process, ensuring more robust and accurate predictions. These data preprocessing steps, driven by the insights gained from EDA, contributed to the optimization of feature representation, reduction of multicollinearity, and management of outliers, ultimately enhancing the performance and interpretability of our machine learning models.

\subsection{Modeling Framework}
The ordered list of $(cs_1-ps_1), (cs_2-ps_2), \ldots, (cs_n-ps_n)$ could be framed as a regression problem. In our work, we employed multiple XGBoost Regressors to tackle this task. One XGBoost model was trained to predict the delta in overall account-level spend, while multiple product-level XGBoost models were trained to predict the delta in product quantity across various product categories. These models were trained using two years of historical account and product bookings data and are currently retrained on a monthly basis. The magnitude of output of the models $(cs_i-ps_i)$ is dependent on the size of the account and its baseline spend. To make accounts with different base sizes and having different ranges of account spend comparable among each other and generate a rank order for prioritization, we normalize the model outputs by passing them onto a log(base spend) based normalization step. Specifically,
\begin{enumerate}
    \item 
    Account spend model: Score range of $(cs_i-ps_i)$ is normalized to $[-100, 100]$.
    \item 
    Product quantity models:  Score range of $(cq_i-pq_i)$ is normalized to $[-10K, 10K]$.
\end{enumerate}

The resulting normalized and rank-ordered list of accounts is then passed to the explanation layer, along with input features, to generate instance-level explanations. Post explanation generation we then feed the scores into the sales rep facing tools and CRM systems. The normalized score ordered list gives sales reps the ability to compare across their account book and decide whether they want to focus on upsell or churn. Many sales reps leverage these scores in a $2\times2$ table with the scores on $x$ axis and renewal target on $y$ axis to identify the segments of accounts such as high spend + high likelihood to grow, high spend + high churn risk, low spend + low churn risk and so on, and devise their sales strategies accordingly. Figure \ref{fig:model_architecture_and_deployment} shows the overall model architecture and deployment.

\subsection{Explanation Generation}

A key thing we learned from a focus group study with sales reps is that the scores alone may not be the most helpful. For sales reps to take action, they need to know the underlying reasons behind these scores, and they also want to double check these reasons with their domain knowledge. Even though some state-of-the-art model interpretation approaches (e.g., LIME\cite{ribeiro2016should}, SHAP\cite{lundberg2017unified}) can help create an important feature list to interpret the ML-model provided scores, the feature names in these lists are often not very intuitive to a non-technical audience. The features also may not be well-organized (e.g., relevant features could be further grouped, redundant features could be removed).

\begin{figure}
\centering
  \includegraphics[width=\columnwidth]{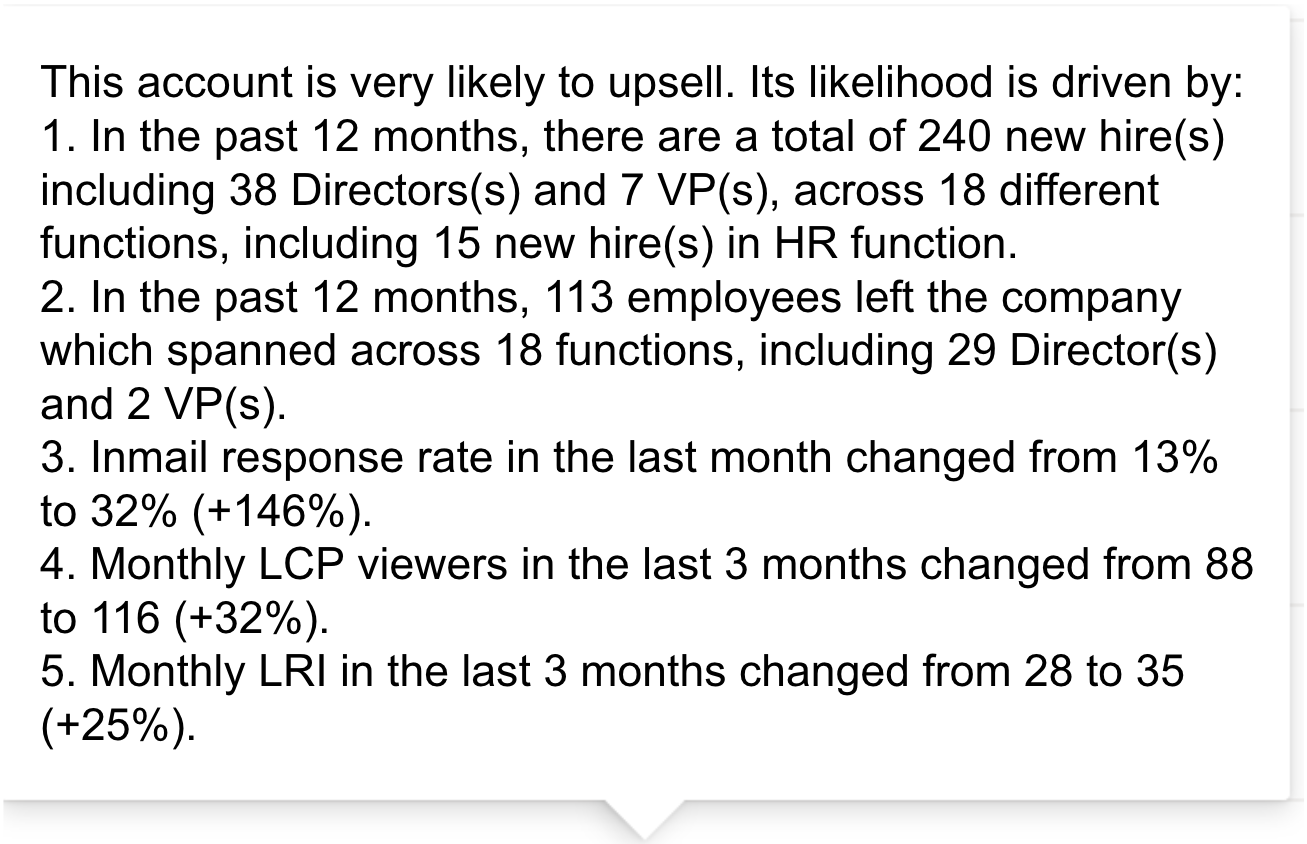}
  \
  \caption{Mocked top narrative insights generated by CrystalCandle for a specific customer in account level upsell/churn prediction.}~\label{fig:narrative_insights_I}
\end{figure}

To deal with the above challenges, we have built and implemented a user-facing model explainer called CrystalCandle\cite{yang2021crystalcandle}, which is a key part of developing transparent and explainable AI systems at LinkedIn. The output of CrystalCandle is a list of top narrative insights for each customer account (shown in Figure \ref{fig:narrative_insights_I}), which reflects the rationale behind the ML-model provided scores. These narrative insights are much more user-friendly, give more support for sales teams to trust the prediction results and better extract meaningful insights.

\begin{figure}
\centering
  \includegraphics[width=\columnwidth]{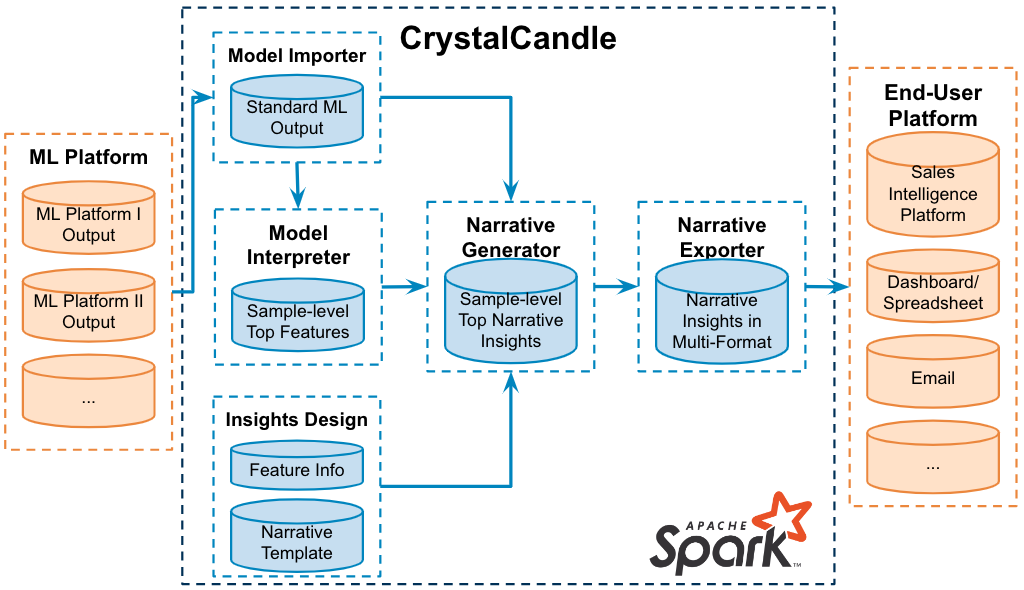}
  \
  \caption{CrystalCandle pipeline.}~\label{fig:crystalcandle}
\end{figure}

Figure \ref{fig:crystalcandle} shows the pipeline of CrystalCandle. CrystalCandle serves as a bridge between the machine learning models (e.g., upsell/churn prediction models) and the end users (e.g., sales reps). The Model Importer extracts the model output from a set of major machine learning platforms (e.g., ProML \cite{young2019proml}), and then in Model Interpreter, we implement model interpretation approaches onto the extracted machine learning model output and generate the important feature list for each sample. The Model Interpreter is compatible with state-of-the-art model interpretation approaches such as SHAP\cite{lundberg2017unified}, LIME\cite{ribeiro2016should}, K-LIME\cite{hall2019introduction}, and FastTreeSHAP\cite{yang2021fast}. We also feed some additional inputs into CrystalCandle at this stage, including the additional feature information and narrative templates. We then conduct narrative template imputation in Narrative Generator and produce top narrative insights for each sample. Finally, we surface these narrative insights onto a variety of end-user platforms via Narrative Exporter.

\subsubsection{Narrative Generator and Insights Design deep dive}

The goal of Narrative Generator is to produce the top narrative insights based on model output and model interpretation results.

\begin{figure*}
\centering
  \includegraphics[width=1.8\columnwidth]{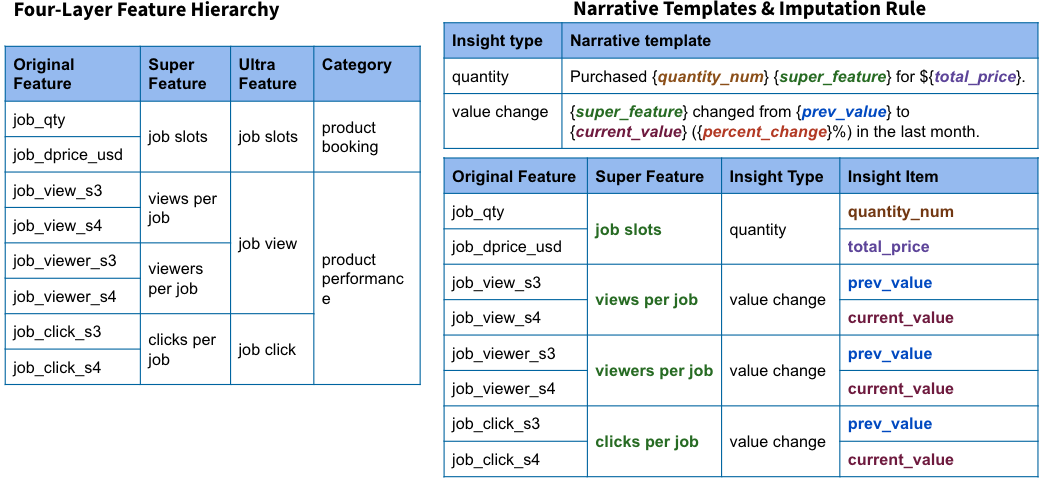}
  \caption{Feature clustering and template imputation in Narrative Generator.}~\label{fig:narrative_generator_I}
\end{figure*}

Figure \ref{fig:narrative_generator_I} shows how we build the feature clustering information file and narrative templates in Narrative Generator. We build the four-layer feature hierarchy as the feature clustering information file. For each original feature, we figure out its higher level features, moving from super feature to the category. We see that the feature descriptions have been naturally incorporated in the super feature names. We also construct a list of narrative templates, where each template is uniquely identified by its insight type. The rule to conduct narrative template imputation is provided by the last table, where one narrative will be constructed for each super feature. The insight item determines the position to impute the feature values into the narrative templates. For example, to construct the narrative for super feature “viewers per job,” we find out its narrative template “value change,” replace the blanks “prev\_value”, “current\_value”, and “super\_feature” with the feature values and super feature name “viewers per job,” and calculate “percent\_change.”

\begin{figure*}
\centering
  \includegraphics[width=1.8\columnwidth]{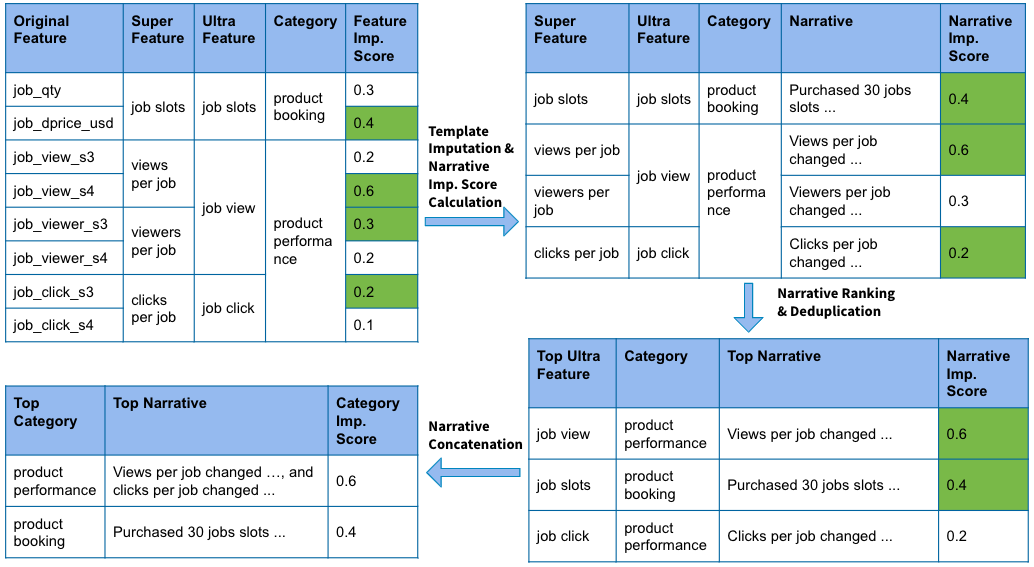}
  \caption{Narrative ranking and deduplication in Narrative Generator.}~\label{fig:narrative_generator_II}
\end{figure*}

We next show how we select top narratives in a scalable way in Figure \ref{fig:narrative_generator_II}. We first append the feature importance scores from Model Interpreter into the feature information table presented in Figure \ref{fig:narrative_generator_I}. During the narrative imputation process, we also calculate the narrative importance score as the largest feature importance score of all the original features appeared in the narrative, and use the narrative importance score to rank all the narratives (heuristics-based reweighting can also be implemented to prioritize actionable narratives). In the meantime, we also conduct narrative deduplication by keeping only the narrative with the largest narrative importance score within each ultra feature. This is in consideration of the fact that narratives under one ultra feature can be highly overlapped. Finally, we conduct narrative concatenation by concatenating narratives within each category; the concatenated top narratives are the final output from the Narrative Generator.

\subsubsection{CrystalCandle implementation details}

\begin{figure}
\centering
  \includegraphics[width=\columnwidth]{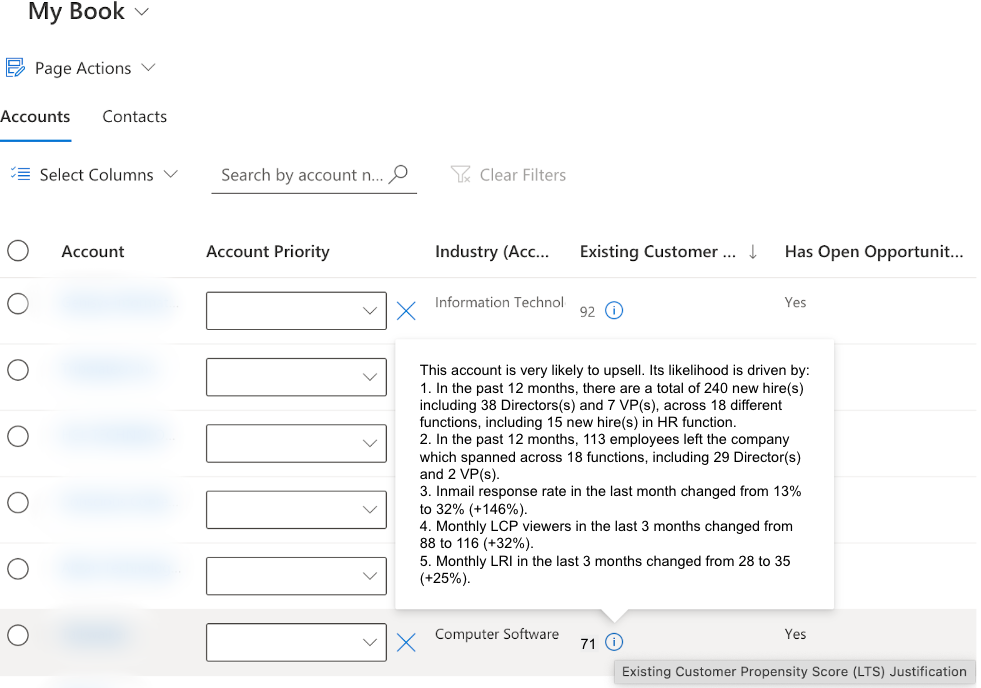}
  \caption{Mocked CrystalCandle output (Customer Propensity Score insights illustrative example) on MyBook.}~\label{fig:narrative_insights_II}
\end{figure}

LinkedIn sales teams use multiple internal sales intelligence platforms. One typical platform, MyBook (embedded in Microsoft Dynamics), aims to help sales reps close deals faster by providing well-organized sales insights and recommendations. Figure \ref{fig:narrative_insights_II} shows one typical output of CrystalCandle on MyBook in Project Account Prioritizer. When a sales rep logs into MyBook, a list of accounts are displayed on the MyBook homepage. The column “Existing Customer Propensity Score (LTS) Justification” shows the upsell/churn propensity score for each account from the predictive models. To learn more about the underlying reasons behind each score, sales reps can hover over the “i” button and a small window with more account details will pop up. In this pop-up window, CrystalCandle provides top narrative insights for each account. Feedback from the sales team has been highly positive: “The predictive models [are] a game changer! The time saved on researching accounts for growth opportunities has been cut down with the data provided in the report which has allowed me to focus on other areas across MyBook.”

\subsection{Measurement}

\subsubsection{Launch with an A/B Test}
The primary objective of the account prioritization engine is to enhance the efficiency of sales reps and drive growth in B2B sales revenue. To evaluate its effectiveness, we conducted an A/B test. However, conducting experiments in the sales domain presents unique challenges, including:
\begin{enumerate}
    \item 
    Low sample size: The sample size is inherently limited due to the number of existing customers and the number of sales reps.
    \item 
    Lagged treatment effect: The sales cycle, from initial customer discussions to opportunity closure, typically spans 3-6 months. Consequently, measuring the impact of account prioritization on sales reps or individual accounts requires a significant amount of time.
    \item 
    Fairness: Randomizing the treatment based on sales reps may result in some reps receiving the account scores while others do not. If these scores prove to be beneficial, it could potentially influence the sales performance and compensation of the reps involved.
\end{enumerate}

\begin{figure}
\centering
  \includegraphics[width=\columnwidth]{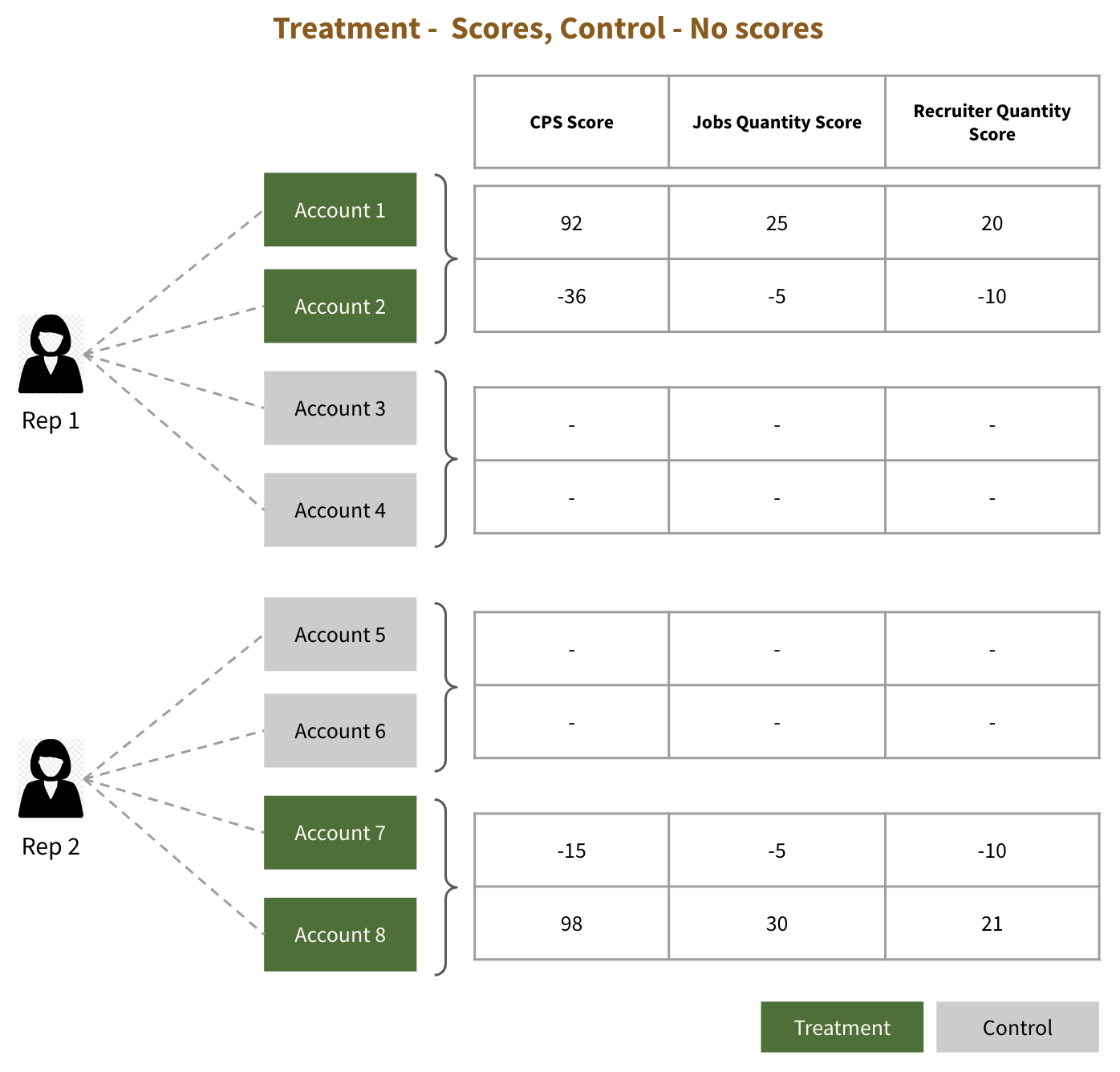}
  \caption{AB test randomization.}~\label{fig:ab_test}
\end{figure}

To address these challenges, we implemented a stratified randomization technique (Figure \ref{fig:ab_test}), wherein randomization occurs at the level of individual sales reps and their respective accounts, with accounts being stratified by matching across variables such as account size, sales segment and region. This approach ensures that each rep observes the model scores for a randomly selected 50\% of their accounts, while the remaining 50\% act as the control group with no scores displayed (Table \ref{tab:ab_test}). In addition, the account stratification ensures similar distribution of accounts in treatment and control. To maintain transparency and fairness, we proactively communicated this change to the reps, emphasizing that even accounts without scores still provide opportunities for customer growth and sales reps can identify these opportunities leveraging their field knowledge. Although this randomization technique may not completely eliminate the placebo effect, it offers the following advantages:
\begin{enumerate}
    \item 
    Ensures control is the status quo, i.e. no scores and sales reps using their intuition and field knowledge to find opportunities.
    \item 
    Offers a fairer alternative to reps as compared to displaying randomized and distorted scores as placebo for the control accounts.
\end{enumerate}

In the A/B test, we measured the difference in Renewal Incremental Growth (RIG, which measures the ratio between renewal actual bookings/expected bookings) between the treatment and control groups. Let $t$ denote treatment and $c$ denote control, $r$ represent the number of reps, and $n$ the number of accounts. The specific metric we measured was the following:
\begin{equation*}
\begin{split}
    &\Delta \text{ in Renewal Incremental Growth} \\
    =&(\text{Renewal Incremental Growth})_t - (\text{Renewal Incremental Growth})_c\\
    =&\frac{1}{\sum_{i\in {r_k}}n_{i,t}}\left[\sum_{i=1}^{r}\sum_{j=1}^{n_{i,t}}(\frac{\text{Renewal Bookings}}{\text{Renewal Target}})_{j,t}\right]\\
    &-\frac{1}{\sum_{i\in {r_k}}n_{i,c}}\left[\sum_{i=1}^{r}\sum_{j=1}^{n_{i,c}}(\frac{\text{Renewal Bookings}}{\text{Renewal Target}})_{j,c}\right],
\end{split}
\end{equation*}
where $n_{i,t}$ and $n_{i,c}$ stand for the number of treatment accounts and the number of control accounts for sales rep $i$ respectively.

\begin{table*}
  \caption{AB test randomization details and measurement results.}
  \label{tab:ab_test}
  \begin{tabular}{l|p{6cm}|p{6cm}}
    \toprule
    Variant & Treatment & Control\\
    \midrule
    Definition & Treatment accounts of reps who visited the data product at least once & Control accounts of reps who visited the data product at least once\\
    \midrule
    Product Experience & Scores with account-level explanations & No scores\\
    \midrule
    Sample size & $\sim$4.5K accounts & $\sim$4.5K accounts\\
    \midrule
    Test Duration & \multicolumn{2}{l}{6 Months} \\
    \midrule
    Success Metric & \multicolumn{2}{l}{Renewal Incremental Growth (RIG)} \\
    \midrule
    Result & \multicolumn{2}{l}{+8.08\% lift in Average RIG of treatment vs control} \\
  \bottomrule
\end{tabular}
\end{table*}

Furthermore, to calculate the Average Treatment Effect on Treated (ATT), we tracked sales reps who viewed the scores at least once, and calculated the difference in RIG between their treatment and control accounts. The experiment yielded positive results, and we observed an +8.08\% lift in RIG for the treatment accounts as compared to control (Table \ref{tab:ab_test}).

\subsubsection{Shifting to Consistent MAUs and Causal Measurement}

In the context of sales optimization, wherein the sales reps self select whether and how frequently they will access the scores, we face the challenge of rep adoption and its downstream effect on model impact. The 8.08\% increase in Renewal Incremental Growth (RIG) identified through the A/B test is applicable only when the sales rep visits the data product where scores are displayed. With constant messaging and initiatives driving adoption, we were able to achieve an impressive 85\% cumulative adoption rate (defined as a one-time view of the scores), but over the course of this data product maturity, we realized the need for measuring the incremental impact of using the scores more frequently as compared to ad hoc usage. Additionally, we observed that the cohort of sales reps viewing the score more frequently had higher RIG than the reps who were ad hoc users. Hence, there was a need to define a retention metric, examine the effect of differentiated usage and understand whether the effect is causal. To do so, we introduced the following definitions:
\begin{enumerate}
    \item 
    Consistent Monthly Active User (Consistent MAU): Sales reps who were MAUs for 4 or more months within a 12-month period.
    \item 
    Infrequent User: Sales reps who were MAUs for 1 to 3 months within a 12-month period.
\end{enumerate}

\begin{table*}
  \caption{MAU based causal measurement details and results.}
  \label{tab:mau_measurement}
  \begin{tabular}{l|p{6cm}|p{6cm}}
    \toprule
    Variant & Treatment & Control\\
    \midrule
    Definition & Infrequent users who turned into Consistent MAUs & Infrequent users who continued to be infrequent users\\
    \midrule
    Sample size & 280 sales reps & 596 sales reps\\
    \midrule
    Study Duration & \multicolumn{2}{p{12cm}}{2 years lookback (6 months pre-treatment for A/A test, 12 months of treatment, 6 months of response collection)} \\
    \midrule
    Success Metric & \multicolumn{2}{l}{Renewal Incremental Growth (RIG)} \\
    \midrule
    Result & \multicolumn{2}{l}{+20.4\% lift in Average RIG of treatment vs control} \\
  \bottomrule
\end{tabular}
\end{table*}

Given the scores were already 100\% ramped, ruling out the possibility of another AB experiment, we conducted a causal measurement study to quantify the incremental impact of transitioning from an infrequent user to a Consistent MAU. Since the effect is directly linked to a sales rep's behavior and extends to all accounts in their portfolio, we needed to conduct the study at the sales rep level rather than the account level. Using a coarsened exact matching (CEM) \cite{iacus2012causal} based matching method, we matched treated units (sales reps who became Consistent MAUs from infrequent users) with control units (sales reps who continued to be infrequent users) based on a set of confounders that affect the treatment and the likelihood of being treated (Table \ref{tab:mau_measurement}). The confounders we controlled for include account size, rep region, sales segment, account spend, rep tenure, macro along with other business changes. In order to make this comparison more robust we:
\begin{enumerate}
    \item 
    Only take the users who converted into a Consistent MAU from an infrequent user during the time adoption activities were carried out.
    \item 
    Apply a A/A test validation, wherein during pre-treatment intervention we check whether the treatment and control cohorts did not have any statistically significant difference in RIG.
    \item 
    Carried out detailed balance and coverage checks, trying to ensure that the coverage of matched accounts in control \& treatment does not drop below 80\% (for generalization of treatment effect), while lowering imbalance between the two groups.
\end{enumerate}

Let's say the matched group of reps are $R = r_1 \cup r_2 \cup \cdots \cup r_k$, then we are measuring,
\begin{equation*}
\begin{split}
    &\Delta \text{ in Renewal Incremental Growth} \\
    =&(\text{Renewal Incremental Growth})_t - (\text{Renewal Incremental Growth})_c\\
    =&\sum_{k=1}^r\frac{1}{w_{r_k}}\bigg\{\frac{1}{\sum_{i\in {r_k}}n_{i,t}}\left[\sum_{i\in r_k}\sum_{j=1}^{n_{i,t}}(\frac{\text{Renewal Bookings}}{\text{Renewal Target}})_{j,t}\right]\\
    &-\frac{1}{\sum_{i\in {r_k}}n_{i,c}}\left[\sum_{i\in r_k}\sum_{j=1}^{n_{i,c}}(\frac{\text{Renewal Bookings}}{\text{Renewal Target}})_{j,c}\right]\bigg\},
\end{split}
\end{equation*}

where $w_{r_k}$ is the CEM weight for matched group $r_k$. We observed a +20.4\% lift in RIG for the sales reps who converted to Consistent MAUs during the treatment period as compared to the sales reps who continued to be infrequent users (Table \ref{tab:mau_measurement}). This further strengthened the need to use the scores more consistently in the sales process and boosted rep adoption.

\section{Conclusion and Future Directions}
Scaling intelligent and value-driven customer outreach for the LinkedIn sales team is a crucial business challenge. The LinkedIn Data teams developed state-of-the-art machine learning models in Account Prioritizer to provide the sales team with account-level information on churn risk and upsell propensity. We further leveraged the user-facing model explainer CrystalCandle to create top narrative insights for each account-level recommendation. CrystalCandle helps sales teams trust modeling results and extract meaningful insights from them. A/B testing results demonstrate that the launch of Account Prioritizer with CrystalCandle interpretation has led to significant revenue improvements. The sales reps adoption (at 85\%) showed strong user intent to use the data product.

Although Account Prioritizer + CrystalCandle offers data-driven guidance on renewal and upsell opportunities, it has room to further develop into a much more agile, personalized and effective recommendation engine with actionable guidance and detailed materials for facilitating customer engagement. For example, additional layers of optimization could be introduced to prioritize the weekly account-related tasks performed by sales reps, such as customer training, sharing of usage reports, conducting product demonstrations, and engaging in budget discussions. By leveraging multitask learning techniques, we can seamlessly integrate the optimization of customer acquisition and existing customer accounts, ensuring a well-balanced approach to optimizing the entire sales funnel. To provide proactive support to sales reps, the system can incorporate dynamic alerts to notify them about upcoming renewal activities, along with offering Next Best Actions (NBAs) to guide their actions. Once the next best action is identified, the recommender can leverage the potential of Generative AI features to generate sales outreach drafts and automate customer reports as part of the pitching material. In essence, this development aims to create a collaborative sales co-pilot that works alongside sales reps to optimize their performance.

\section*{Acknowledgments}
We would like to express our sincere appreciation to our colleagues at LinkedIn for their contributions in making this data product a reality, including Liyang Zhao, Adam Chard, Das Apparsamy, Jacqueline Rivas, Jessica Li, Diana Negoescu, Saad Eddin Al Orjany, Yu Liu, Wenrong Zeng, Samba Njie, Xia Hong, Katie Stohs, Rodrigo Aramayo, Kunal Chopra, Guy Berger, Brian Weller, Farhan Syed, Mark Lobosco, Parvez Ahammad, Faisal Farooq, Rehan Khan and Ya Xu. We particularly thank Jiang Zhu and Stephanie Sorenson for their helpful comments and feedback.

\bibliographystyle{ACM-Reference-Format}
\bibliography{ap_reference}




\end{document}